\documentclass[10pt,a4paper]{article}
\usepackage[utf8]{inputenc}
\usepackage{bm}
\usepackage{amsmath}
\usepackage{multicol}
\usepackage[T1]{fontenc}
\usepackage[top=1.5cm, bottom=2cm, left=2.5cm, right=2.5cm]{geometry}
\usepackage{amssymb,amsthm} 
\usepackage{fancyhdr} 
\usepackage{titlesec}
\usepackage{textpos}
\usepackage{mdwlist}
\usepackage{color}
\usepackage{cancel}
\usepackage{graphicx}
\usepackage{pgf,tikz}
\usetikzlibrary{arrows}
\usepackage{tabularx}
\usepackage{setspace}

\titleformat{\section}{\large\bfseries}{\thesection}{1em}{}
\titleformat{\subsection}{\bfseries}{\thesubsection}{1em}{}
\titlespacing{\section}{0pt}{*1.5}{0em}
\titlespacing{\subsection}{0pt}{*1.5}{0em}

\newtheorem{lemma}{Lemma}

\newcommand{\argmin}{\operatornamewithlimits{argmin}}
\newcommand{\diag}[1]{\operatorname{diag}(#1)}
\newcommand{\trace}[1]{\operatorname{trace}\left(#1\right)}
\newcommand{\ex}[1]   {\operatorname{E}\!\left[ #1 \right]}
\newcommand{\exc}[2] {\operatorname{E}\!\left[ #1  \left\vert\vphantom{#1#2}\right.\! #2 \right]}
\newcommand{\exs}[1] {\exc{#1}{s}}
\newcommand{\soll}[4] { \operatorname{solve}_{#1}(#2,#3,#4) }
\newcommand{\sollo}[3] { \soll{L_1}{#1}{#2}{#3} }
\newcommand{\sollt}[3] { \soll{L_2}{#1}{#2}{#3} }
\DeclareMathSymbol{\sminus}{\mathord}{operators}{"2D}

\newcommand{\valfa}[1] { \tilde{V} \left(#1 \right) }
\newcommand{\qL}[0] {I - \gamma P}

\newcommand{\ones} { 1_{\Box} }

\newcommand{\qtiitle}[0] {Properties of the Least Squares Temporal Difference learning algorithm}
\newcommand{\qtiitleshort}[0] {Properties of the LSTD algorithm}
\newcommand{\qauthor}[0] {Kamil Ciosek}

\fancypagestyle{plain}{
\fancyhf{} 
\fancyfoot[C]{\thepage} 
}

\begin{document}

\thispagestyle{plain}

\begin{center}
{\large\bfseries \qtiitle} \\[1em]
\qauthor
\end{center}

\hspace{2em}
\begin{abstract}
This paper focuses on policy evaluation using the well-known Least Squares Temporal Differences (LSTD) algorithm. We give several alternative ways of looking at the algorithm: the operator-theory approach via the Galerkin method, the statistical approach via instrumental variables as well as the limit of the TD iteration. Further, we give a geometric view of the algorithm as an oblique projection. Moreover, we compare the optimization problem solved by LSTD as compared to Bellman Residual Minimization (BRM). We also treat the modification of LSTD for the case of episodic Markov Reward Processes.  
\end{abstract}
\hspace{2em}

The main practical problem that the LSTD algorithm solves is such: we are given a feed of data from a stochastic system, consisting of a state description in terms of features and of rewards. The task is to construct an abstraction that maps from states to values of states, where the value is defined as the discounted sum of future rewards. We will show that for LSTD, this abstraction is a linear model. For example, the system may describe a chess game, the features of state may describe what pieces the players have while the reward signal corresponds to wither winning or losing the game. The value signal will then correspond to the value of having each particular piece. Note that this is not a general constant but may depend on the way the individual players play the game, for example the values may be different for humans than for computer players. It is well-known that the value function of a given policy can be expressed as $V = (I - \gamma P)^{-1} R$. The LSTD algorithm can be thought as a way of computing the value of this function approximately. The motivation for why the approximation is often necessary is threefold. First, we may not have access to the states directly, just to functions $\phi$ of state. Second, the number of states $n$ is often computationally intractable. Third, even if $n$ is tractable, there is the problem of statistical tractability -- the number of samples needed to accurately estimate transition matrices $n \times n$ is often completely prohibitive. 

Associated with our problem setting is the question whether the value function is interesting in its own right, or whether we only need it to adjust the future behaviour of some aspect of the environment we can control (i.e.\, in our chess example. make a move). We believe that there is large scope of systems (for instance expert systems) where the focus will be on gaining insight into the behaviour of the stochastic system, but the decisions about whether or how to act will still be made manually by human controllers, on the basis of the value-function information. These are the cases where algorithms like LSTD are the most directly applicable. On the other side of the spectrum, there will also of course be situations where the value function estimate is used as a tool to automatically generate the best action on the part of the agent -- such systems may also use value-function estimation algorithms of the kind of LSTD to operate within the policy iteration framework.

\section{Prior Work on LSTD}
An exhaustive introduction to least-squares methods for Reinforcement Learning is provided in chapter 6 of Bertsekas' monograph \cite{bertsekas-vol2}. The LSTD algorithm was introduced in the paper by Bradtke and Barto \cite{lstd-bb}. Boyan later extended to the case with eligibility traces \cite{boyan2002technical}, wherean additional  parameter $\lambda$ controls how far back the updated are influenced by previous states. The connection between LSTD and LSPE, as well as a clean-cut proof that the on-line version of LSTD converges, was given by Nedi{\'c} and Bertsekas \cite{nedic2003least}. The seminal paper \cite{tsitsiklis1997analysis} by Tsitsiklis and Van Roy provided an explicit connection between the fix-point of the iterative TD algorithm and the LSTD solution, while also formally proving that the TD algorithm for policy evaluation converges. The paper \cite{baird1995residual}, described the Bellman Residual Minimization procedure as an alternative to TD. Antos' paper \cite{antosbrm} provided an extensive comparison on the similarities and differences between LSTD and Bellman Residual Minimization (BRM). Parr's paper \cite{lagoudakis2003least} introduced the LSPI algorithm as a principled way to combine LSTD with control.  The paper by Munos \cite{munos2003error} introduced bounds for policy iteration with linear function approximation, albeit under strong assumptions. Scherrer provided \cite{scherrer-brm} the geometric interpretation of LSTD as an oblique projection, in the context analysing the differences between LSTD and BRM. The paper \cite{keller2006automatic} represents an early approach to automatically constructing features for RL algorithms, including LSTD. Schoknecht gave \cite{schoknecht2002optimality} an interpretation of LSTD and other algorithms in terms of a projection with respect to a certain inner product. Choi and Van Roy \cite{choi2006generalized} discuss the similarities between LSTD and a version of the Kalman filter. There exist various approaches in literature to how LSTD can be regularized, none of which can be conclusively claimed to outperform the others. These include the L1 approaches of \cite{loth2007sparse} and  \cite{kolter-ng-reg} and the nested approach of  \cite{hoffman-lstdreg}. These approaches differ not just in the what regularization term in used, but they solve different optimization problems (we will discuss this in section \ref{sec-reg}).

\section{Definition of LSTD} \label{def-lstd}
\subsection{Notation}
The LSTD algorithm finds the value function of a finite-state Markov Reward Process (MRP). The MRP is fixed, i.e.\ we only consider the on-policy setting. We only have access to linear features of states and to the obtained rewards. More formally, denote as $P$ the transition matrix of the MRP. For each state $s$ we have a feature row-vector $\phi$. The feature design matrix $\Phi$ gives the features of all states of the MRP, row-wise, where we assume that $\Phi$ has independent columns. We use the vector $R$, the $i$-th element of which contains mean reward obtained while leaving the state $i$. We use $\xi$ to denote a left eigenvector of $P$ corresponding to eigenvalue one. Note that if the chain has a stationary distribution, it will correspond to such an eigenvector, but we do not require it. We will assume that the chain only has one recurrent class since the case where we have many classes complicates the notation without contributing to the main argument (in practice, we can typically assume there is one class if we do enough exploration). We also introduce the matrix $\Xi = \diag{\xi}$. We now define expectations of functions of the Markov process in terms of weighted averages. For example the expectation of $\phi^\top \phi$, is defined by $\ex{\phi^\top \phi} = \Phi^\top\Xi\Phi$, and similarly for other functions. By this we mean that if $P$ is ergodic, it is legitimate to consider the above quantity an expectation corresponding to long-time average by the standard ergodic theorem for Markov chains. But in our application it is convenient to be more general and allow for periodicity, i.e.\ the diagonal of $\Xi$ may not be a stationary distribution, but the expression still matches the long-time average. We use subscripts do denote two-step sampling, for example $\phi_s'$ denotes the fact that we first sample a state, then the successor state and obtain the feature of that successor state. When we write an expectation w.r.t.\ such a variable, for example $\ex{r_s^2}$, the distribution we mean for $r$ is $ \sum_{s=1}^S p(r|s) \xi_{s}$. Part of our present derivations depends on treating some qualities as random variables; we use small letters to denote them, for instance $s$ denotes state and $\phi$ denotes feature. Once we have obtained samples from our process, we store them in matrices $\hat{\Phi}$ and $\hat{r}$, whose $i$-th rows correspond to, respectively, the state feature vector and reward obtained at time $i$. Observe the difference between $\Phi$ and $\hat{\Phi}$ -- in the first one, each state is represented once, in the second one the number of rows corresponds to the trajectory taken in the MRP and repetitions are possible. The value function is discounted with the factor $0 < \gamma < 1$. Moreover, we introduce the square matrix $D$ which has ones on the main diagonal and $-\gamma$ on the diagonal above it. It is the sample based equivalent to the operator $I-\gamma P$. 
\[
D = \left [ \begin{array}{cccc} 1 & -\gamma &  &  \\  &  \ddots & \ddots  & \\  &    & 1 & -\gamma \\ &    &  & 1 \\ \end{array} \right ]
\]

\subsection{The linear dynamical system approach}
\label{sec-lds}
The derivation given in this section is based on \cite{Parr2008}. We begin by constructing a MRP which lives in the space of features instead of our original state space. We limit ourselves to the class of linear dynamical systems. We need to define the matrix $\tilde{P}$ and the vector $\tilde{R}$, so that a transition from $\phi$ to $\phi'$ (row vectors) is modelled by $\phi\tilde{P} = \phi'$, and the reward we expect at $\phi$ is modelled by $\phi\tilde{R} = r$. Now we look for the values for $\tilde{P}$ and $\tilde{R}$ that model our system dynamics. We have that $\Phi\tilde{P}$ should be approximately equal to $P\Phi$ and $\Phi\tilde{R}$ to $r$. We weigh states by $\Xi$, giving the following optimization problems.
\begin{gather}
\tilde{P} = \argmin_{\tilde{P}} \|\Phi \tilde{P} - P \Phi\|_\Xi = \argmin_{\tilde{P}} \trace{(\Phi \tilde{P} - P \Phi)^\top \Xi (\Phi \tilde{P} - P \Phi)} \notag \\ 
\tilde{R} = \argmin_{\tilde{R}}\ \|\Phi \tilde{R} - R\|_\Xi = \argmin_{\tilde{R}}\; (\Phi \tilde{R} - R)^\top \Xi (\Phi \tilde{R} - R)
\label{e-ldsopt}
\end{gather}
These optimization problems correspond to ordinary least squares (generalized to matrices in case of $\tilde{P}$) and the solutions are obtained by weighted projection: $\Phi \tilde{P} = \Pi P \Phi$ and $\Phi \tilde{R} = \Pi R$, where the projection matrix is defined as $\Pi = \Phi (\Phi^\top \Xi \Phi)^{-1} \Phi^\top \Xi$ and the matrix $\Phi$ cancels with the one in the projection, since it is full column rank. Now consider a feature vector $\phi$. In the new approximate MRP, we can compute the value function exactly (i.e.\ all the approximation has already taken place when we constructed the matrix $\tilde{P}$ and vector $\tilde{R}$). The true value function associated with it is the expected discounted future reward, and is expressed as follows.
\begin{gather}
\label{eq-lstd-lds}
\phi \underbrace{\sum_{i=0}^\infty (\gamma \tilde{P})^i \tilde{R}}_{w}  = \phi \underbrace{(I - \gamma \tilde{P})^{-1}  \tilde{R}}_{w}
\end{gather}
In the above, the last equality is the well known von Neumann telescoping sum argument. We thus have the equation $(I - \gamma \tilde{P}) w = \tilde{R}$. In the above, we assumed that the series $\sum_{i=0}^\infty (\gamma \tilde{P})^i \tilde{R}$ converges. We show a stronger condition, namely that the series $\sum_{i=0}^\infty (\gamma \tilde{P})^i$ converges, which is the same as saying that $\gamma \tilde{P}$ is a contraction in some norm. 

This follows from the following reasoning. Consider first the case when we have $\Xi > 0$. We know that $\Pi P$ is a contraction in the norm weighted by $\Xi$ i.e. $\| \Pi P\|_\Xi = \| \Xi^{\frac12} \Pi P \Xi^{-\frac12} \|_2 \leq 1$ (see for example \cite{bertsekasc}, proposition 6.3.1). Therefore the spectral radius of $\Pi P$ is bounded by one. Define the matrix $\Pi^\sminus = (\Phi^\top \Xi \Phi)^{-1} \Phi^\top \Xi$, so that we have $\Pi P = \Phi ( \Pi^\sminus P)$ and $\tilde{P} = (\Pi^\sminus P) \Phi$. Using the assumption that $\Phi$ has independent columns, it is easy to see that if $v$ is an eigenvector of  $\Pi^\sminus P \Phi$ then $\Phi v$ is an eigenvector of $\Phi ( \Pi^\sminus P)$ with the same eigenvalue. Hence all eigenvalues of $\tilde{P}$ are also eigenvalues of $\Pi P$ and $\rho(\tilde{P}) \leq 1$. Now if we have some zero entries on the diagonal of $\Xi$, our results follows by a continuity argument. Thus we have the result for the general case.

Note that in the above proof we used the fact that $\Xi$ has on the diagonal is a left eigenvector of $P$ corresponding to eigenvalue one. If $\Xi$ used for the projection were an \emph{arbitrary} distribution, then the matrix $\tilde{P}$ would in general have spectrum beyond the unit circle. For example, consider the following.

\[
P = \left [ \begin{array}{cccc} 1 & 0 & 0 & 0  \\ 1 & 0 & 0 & 0 \\ 0 & 1 & 0 & 0 \\ 0 & 0 & 1 & 0 \\ \end{array} \right ] \quad
\Phi = \left [ \begin{array}{cc} 1 & 1  \\ 1 & 0 \\ 0 & 1 \\ 0 & 0 \\ \end{array} \right ]
\]

Now, if we assume a uniform distribution $\Xi$, we obtain the following matrix, the spectral radius of which is more than one.
\[
\tilde{P} = \frac13 \left [ \begin{array}{cc} 2 & 3  \\ 2 & 0 \end{array} \right ]
\]

\subsection{Interpretation in terms of expectations}
We now give the interpretation of the matrices $\tilde{P}$ and $\tilde{R}$ in terms of expectations, which are useful when constructing sample-based versions of the algorithm. 
\[\tilde{P} = \textstyle (\Phi^\top \Xi \Phi)^{-1} \Phi^\top \Xi P \Phi = \ex{\phi_s^\top \phi_s}^{-1}\ex{\phi_s^\top \phi_s'}\]
\[\tilde{R} = \textstyle (\Phi^\top \Xi \Phi)^{-1} \Phi^\top \Xi R = \ex{\phi_s^\top \phi_s}^{-1}\ex{\phi_s^\top r}\]

We also can construct sample-based variants of the matrices $\tilde{P}$ and $\tilde{R}$ (call them $\hat{P}$ and $\hat{R}$ respectively) and still obtain essentially the same algorithm. Let us adopt the following definitions.
\[ \hat{\tilde{P}} = \argmin_{\hat{\tilde{P}}} \|\hat{\Phi} \hat{\tilde{P}} - N \hat{\Phi} \| = (\hat{\Phi}^\top \hat{\Phi})^{-1} (\hat{\Phi}^\top N \hat{\Phi})\]
\[ \hat{\tilde{R}} = \argmin_{\hat{\tilde{R}}} \|\hat{\Phi} \hat{\tilde{R}} - \hat{R} \| = (\hat{\Phi}^\top \hat{\Phi})^{-1} (\hat{\Phi}^\top \hat{R}) \]
In the above, we denote by $N$ the matrix which has ones above the main diagonal and zeros elsewhere.
\[
N = \left [ \begin{array}{cccc} 0 & 1&  &  \\  &  \ddots & \ddots  & \\  &    & 0 & 1 \\ &    &  & 0 \\ \end{array} \right ]
\]
We note that the required inverse exists by the assumption that $\Phi^\top \Xi \Phi$ is invertible (we also implicitly assume that we have enough samples). Furthermore, we note that the transition model and the reward model are independent, i.e. our solution is also applicable to the setting where we have a single transition matrix $P$, but instead of just one reward we have many tasks, each of which with a different reward \cite{gc}. We note that in this setting, while we still have to learn $\tilde{R}$ for each task separately, it is worthwhile to learn $\tilde{P}$ using training data from \emph{all} tasks.

\section{Other Ways to obtain LSTD}
We begin with the Bellman equation, which defines the true value function: $V(s) = \exs {r + \gamma V(s') } = \exs {r} + \gamma \exs{V(s') }$. It can be rewritten in matrix form as $V = (I - \gamma P)^{-1} R$. We exploit a linear architecture: i.e. we seek to approximate the true value function $V(\cdot)$ with the function $\valfa{s} = w^\top{\phi}$, which is linear in $w$. We will briefly discuss two possibilities for how to choose an appropriate $\valfa{\cdot}$ within the linear class of functions. The obvious thing would be to define $\bar{V} = \Pi V = \Pi (I - \gamma P)^{-1} R$, where $\Pi = \Phi(\Phi^\top \Xi \Phi)^{-1} \Phi^\top \Xi$ where we assume that the inverse exists. This formula guarantees that the distance from $\bar{V}$ to $V$ is minimal in the elliptic norm weighted by $\Xi$. The problem with this approach is that it is not known how to efficiently compute a useful estimate of the projected value function from samples.\footnote{One algorithm that can do that in the limit of infinitely many samples is Least-Squares Monte-Carlo. It is, however, prone to high variance in the estimate for small sample sizes.} Therefore we need a different approximation. We call it $\valfa{\cdot}$. It comes through the equation $\tilde{V}D = \Pi T \tilde{V}$, where we look for the fixpoint of the operator $\Pi T$ instead of the Bellman operator $T$, where $Tx = R + \gamma P \Phi x$.  Our choice of $\tilde{V}$ will be motivated further later (in particular, see equation \ref{efixpoint}).

Now the question, of course, is about the relation between our approximation $\tilde{V}$ and the projection of the true value function $\bar{V}$, as we have defined it in the previous section. We now state without proof the relation between the two estimates developed in \cite{yu-linarchbound} (see their references for prior work).
\begin{align}
V - \tilde{V} = (I - \gamma \Pi P)^{-1} (V - \bar{V}) \label{linarch}
\end{align}
This can be used to obtain the following bound, which does not require us to estimate the matrices $\Pi$ or $P$ (see \cite{yu-linarchbound} for proof and for sharper bounds): $\|V -\tilde{ V}\|_{\Xi} \leq (1 - \gamma^2)^{-1/2} \|V - \bar{V}\|_{\Xi}$. 

We see from this that one example where the approximation of equation \ref{eb} is appropriate is when we have substantial discounting -- in that case, if the linear framework is good at all, i.e. if the projection $\bar{V} = \Pi V$ is close to the true value function, then so will be our approximation. We emphasise here that our derivation is for the case where there is one recursive class in the MRP. If there are other classes, this bound tells us nothing about them (i.e. using this bound only, we have to accept the value function at the states belonging to them may be arbitrarily off the mark). 

In the subsequent sections, we will describe various seemingly different approaches to computing $\valfa{\cdot}$ from samples, which however all lead to the same formula for the solution we have already seen in section \ref{sec-lds}. In order to derive our algorithm, we make two assumptions. First, we assume the following. We call this the \emph{feature independence assumption}.
\begin{gather}
\label{assm-1}
\ex{\phi^\top \phi} \text{is full rank}
\end{gather}

This implies that the features are linearly independent (i.e. $\Phi$ is of full column rank) but the statement is stronger in that it concerns \emph{both} the features and the transition dynamics of the MRP, and means that the parts of the features corresponding to states visited with nonzero probability are independent. We note that this implies that the matrix $\ex{\phi^\top (\phi - \gamma \phi')}$ is also full rank -- we discuss why this implication holds in appendix \ref{sasm}. We will also use this to claim the invertability of $\hat{\Phi}^\top D \hat{\Phi}$ without further comment (i.e. we assume we have enough samples). Also, we assume that the mean of the reward process exists.
\begin{gather}
\label{assm-2}
\ex{r_s} < \infty 
\end{gather}
 
The second assumption is rarely a problem because in typical applications the reward is bounded by some constant.

To summarize the description, we restate the fundamental conditions for LSTD to yield good value estimates: (1) the linear architecture itself needs to match the problem and the set of features needs to be set right, that is $V$ must be close to $\bar{V}$, (2) the approximation $\tilde{V}$ needs to be good, for example through discounting and finally (3) the sample based approximation $\hat{w}$ to $w$ must also be good (in the following sections we define a consistent estimator for $w$, i.e. a way to compute $\hat{w}$, so that the value function computed from a sample trajectory approaches $\tilde{V}$ for the recursive states in the class corresponding to that trajectory as the length of the trajectory goes to infinity). 

\subsection{Derivation by the Galerkin method}
\label{secgalerkin}
That LSTD corresponds to a special case of the Galerkin argument has been implicitly realized for some time, and formally stated in \cite{bertsekas2011temporal}, on which this section is based. The general idea of the Galerkin method is to approximate the fixed point of $T$, $T x^\star = x^\star$. We have $x^\star = \argmin_x \| Tx^\star - x\|$. We introduce the approximation by considering points from within the column space of $\Phi$, so that our approximate fixpoint satisfies $\tilde{x}^\star \in \text{Range}(\Phi)$, yielding $\tilde{x}^\star = \argmin_{x \in \text{Range}(\Phi)} \| T\tilde{x}^\star - x \|$, which is equivalent to the following, after substituting $\Phi y^\star$ for $\tilde{x}^\star$ and $\Phi y$ for $x$ and using the semi-norm weighted by $\Xi$. 
\begin{align}
\Phi y^\star = \argmin_{y} \| T \Phi y^\star - \Phi y \|_\Xi
\label{eoptgalerkin}
\end{align}
Now, for our semi-norm with the corresponding projection operator $\Pi$, this has an analytic solution: $\Phi y^\star = \Pi (T(\Phi y^\star))$. Now, in our case, $\Pi = \Phi(\Phi^\top \Xi \Phi)^{-1} \Phi^\top \Xi$ where we note that the inverse is well-defined by assumption \ref{assm-1} and the evaluation of the operator $T$ at ${\Phi}w$ becomes $R + \gamma {P \Phi}w$ with $w$ assuming the role of $y^\star$. Now we solve the following.
\begin{align}
{\Phi}w = \Pi ( \underbrace{ R + \gamma {P \Phi}w }_{T{\Phi}w})
\label{efixpoint}
\end{align}
This can be transformed in the following way.
\begin{align}
\label{egalerkin}
\bcancel{\Phi} {(I - \gamma (\Phi^\top \Xi \Phi)^{-1} \Phi^\top \Xi P \Phi))}w = \bcancel{\Phi} (\Phi^\top \Xi \Phi)^{-1} \Phi^\top \Xi R
\end{align}
In the above, we can cancel out the terms $\Phi$, because by assumption \ref{assm-1}, $\Phi$ has to be of full column rank. We then multiply both sides by $(\Phi^\top \Xi \Phi)$, to obtain ${\left((\Phi^\top \Xi \Phi) - \gamma \Phi^\top \Xi P \Phi \right)}w = \Phi^\top \Xi R$, which leads to the following. 
\begin{align}
w = \left(\Phi^\top \Xi \Phi - \gamma \Phi^\top \Xi P \Phi \right)^{-1} \Phi^\top \Xi R
\label{galerkinresult}
\end{align}
This is the same as the expression we will obtain in the instrumental variable section. We also see that equation \ref{egalerkin} is the same as the formula obtained from the linear dynamical system approach in equation \ref{eq-lstd-lds} when we plug in the computed values of $\tilde{P}$ and $\tilde{R}$.  Thus we have obtained the same estimator.

\subsection{Derivation by instrumental variables}
\label{sec-iv}
Again, we begin with the Bellman equation, which defines the true value function: $V(s) = \exs {r + \gamma V(s') } = \exs {r} + \gamma \exs{V(s') }$. We will first obtain a statistical model that expresses the properties of the approximation $\valfa{\cdot}$. By solving the Bellman equation directly in the linear approximation regime, we obtain the following equation.
\begin{align}
{\phi}w = \valfa{s} = \exs {r} + \gamma \exs{\valfa{s'}} - e_{s} = \exs {r} + \gamma {\exs{\phi'} }w - e_{s} \label{eb}
\end{align}
We note that in the above, we use the convention that $w$ is a column vector while the features are row vectors. This convention minimizes the number of transposes we have to write. Note that we had to introduce the TD error vector $e = [e_{s_1},\cdots,e_{s_n}]^\top = T{\Phi}w - {\Phi}w $ and the corresponding random variable $e_{s}$ (i.e.\ the error is a deterministic function of the current state, which is random), since the sum of the reward vector $R$ and the expected feature vector ${\exs{\phi'} }w$ may not be in the feature space (i.e.\ the column space of $\Phi$). It can be verified using equation \ref{galerkinresult} that, the error terms satisfy $e\Xi\Phi = 0$, i.e.\ it is orthogonal to the feature space (indeed it can be seen after a brief manipulation that the condition $e\Xi\Phi = 0$ is \emph{equivalent} to the formula \ref{galerkinresult} -- we will do this in section \ref{td-iterative}), and that consequently we have the following.
\begin{gather}
\label{f-perror}
\Pi e = 0
\end{gather}
This is not a derivation from first principles, since we had to use an external argument to verify that $e\Xi\Phi = 0$ (which is equivalent to assuming that the TD error vanishes in expectation). But given the model of equation \ref{eb} it is nonetheless instructive to look at the mechanics of how the derivation works because this is the first one to have been proposed for LSTD. 

We now accept equation \ref{eb} as a given and give a statistical derivation as provided in the original LSTD paper \cite{lstd-bb}, based on methods described in \cite{wldr-econ}. Now, because we do not observe the expectations $\exs{\gamma V(s') }$ and $\exs {r}$ in equation \ref{eb}, but merely samples of $\phi$ and $\phi'$ we model the residue wrt.\ the expected value as noise, yielding the probabilistic model $r_s = \exs{r} + \eta_s$, where we use assumption given by equation \ref{assm-2}, and $\phi_s' = \exs{\phi' } + \varepsilon_s$. Note that by definition $\exs{\eta_s} = 0$. Observe that this implies the following by the law of iterated expectation (LIE).
\begin{gather}
	 \label{f-noise} 
	 \exc{\eta_s}{\phi} = \exc{\exs{\eta_s}}{\phi} = 0
\end{gather} 
Analogously, we have the following.
\begin{gather}
	\label{f-featurenoise} 
	\exc{\varepsilon_s}{\phi} = 0
\end{gather}
Thus we can rewrite equation \ref{eb} to obtain the following.
\begin{align}
{\phi}w = r_s + \gamma {\phi_s'}w - \gamma {\varepsilon_s}w - \eta_s - e_{s} \quad \text{or} \quad r_s =  {(\phi - \gamma \phi_s')}w + 
\underbrace{\gamma {\varepsilon_s}w + \eta_s}_{\zeta_s} + \; {e_{s}}
\label{ebn}
\end{align}
Now, we cannot use traditional least-squares to solve this, since the expression $\zeta_s = \gamma {\varepsilon_s}w + \eta_s$ may be, in general, correlated\footnote{Indeed, we have $\ex{\phi_s'^\top \eta_s} = 0$, $\ex{\phi^\top \varepsilon_s} = 0$ and $\ex{\phi^\top \eta_s} = 0$ as shown later in the text; but $\ex{\phi_s'^\top \varepsilon_s} = \ex{\phi_s'^\top \phi_s'} - \ex{\phi_s'^\top \exs{\phi_s'}} = \Phi^\top \Xi \Phi - \Phi^\top P^\top \Xi P \Phi$, where the last term does not vanish in general.} with $\phi - \gamma \phi_s'$, so will be the projection error term $e$ and the two correlations will not cancel in general. Therefore the noise term $\zeta_s - e_{s}$ may be correlated with with $\phi - \gamma \phi_s'$. Also, $\exc{e_{s}}{s}$ is not necessarily zero. But ordinary least squares (OLS) requires that noise be uncorrelated with input variables and that it have mean zero to yield consistent estimates. However, there is still a way to obtain a good estimate. More formally, we first need to establish the following properties. First, we have $\ex{\phi^\top \eta_s} = \ex{\exc{\phi^\top \eta_s}{\phi}} = \ex{\phi^\top\exc{\eta_s}{\phi}} = 0$, where the first equality follows from LIE and the second from fact \ref{f-noise}. By the same reasoning, we have $\ex{\phi^\top \varepsilon_s} = 0$ from fact \ref{f-featurenoise}. With these two properties, we can now multiply both sides of equation \ref{ebn} by $\phi^\top$, which we for this purpose call an \emph{instrumental variable}, and then take expectation, so as to make the noise terms vanish. We also have $\ex{\phi^\top e_{s}} = 0$ by fact \ref{f-perror}. This results in the following.
\begin{align}
\ex{\phi^\top r_s} = {\ex{\phi^\top (\phi - \gamma \phi_s')}}w + \underbrace{\gamma {\ex{\phi^\top \varepsilon_s}}w + \ex{\phi^\top \eta_s} - \ex{\phi^\top e_{s}}}_{ =\;0} 
\label{ebnp}
\end{align}
Now because we know by assumption \ref{assm-1} (see section \ref{sasm} of the appendix for a detailed proof) that $\ex{\phi^\top (\phi - \gamma \phi_s')}$ is invertible, the estimator $w$ is given by the following. 
\begin{align}
w &= \ex{\phi^\top (\phi - \gamma \phi_s')}^{-1} \ex{\phi^\top r_s} = \left(\Phi^\top\Xi(I - \gamma P)\Phi\right)^{-1} \Phi^\top \Xi R \quad \text{or} \nonumber \\ \hat{w} &= (\hat{\Phi}^\top D \hat{\Phi})^{-1} \hat{\Phi}^\top \hat{r}
\label{s-lstd}
\end{align}
This finishes the formal derivation. We will now give two different intuitive interpretations to the instrumental variable method. First, consider the sample equivalent of equation \ref{ebn}, which we now rewrite in matrix notation $\hat{r} = {D\hat{\Phi}}\hat{w} + \hat{\zeta} - \hat{e}$, where by $\hat{\zeta}$ we denote the vector containing the noise terms for each individual sample and by $\hat{e}$ the sample values of the random variable $e_{s}$. Now, as described above, we cannot solve it by OLS because of the correlation between the noise and $D\hat{\Phi}$. So we `fix' $D\hat{\Phi}$ by projecting it onto the feature space (i.e.\ the column space of $\hat{\Phi}$), since we know that noise is uncorrelated with features. We introduce the projection operator $\hat{\Pi} = \hat{\Phi}(\hat{\Phi}^\top \hat{\Phi})^{-1} \hat{\Phi}^\top$, where we note that the inverse exists by assumption  given we have enough samples. Now our equation becomes the following.
\begin{align}
\hat{\Pi} \hat{r} = \hat{\Pi} {D\hat{\Phi}}\hat{w} +  \!\!\!\!\!\!\!\!\!\underbrace{\hat{\Pi} \hat{\zeta}}_{\rightarrow 0 \text{ as } N \rightarrow \infty} \!\!\!\!\!\!\!\!\! - \hat{\Pi} \hat{e} \quad \text{or} \quad
\cancel{\hat{\Phi}(\hat{\Phi}^\top \hat{\Phi})^{-1}} \hat{\Phi}^\top \hat{r} = \cancel{\hat{\Phi}(\hat{\Phi}^\top \hat{\Phi})^{-1}} \hat{\Phi}^\top {D\hat{\Phi}}\hat{w}
\label{s-eb}
\end{align}
In the above, we can cancel the terms because $\hat{\Phi}$ has, by assumption , independent columns if we have enough samples. This leads to the same estimator that we derived above. This interpretation is known in econometric literature as two-stage least squares (2SLS), because we solve two linear systems: first we project $D\hat{\Phi}$ on the subspace of features and then we solve the resulting modified equation. In this context we stress that we would get the same solution if we only applied the projection on the right-hand side, e.g.\ $\hat{r} = \hat{\Pi} {D\hat{\Phi}}\hat{w}$ -- this can be seen by noticing that the choice of $\hat{w}$ in this equation is unaffected by any component of $\hat{r}$ orthogonal to the feature space. We also see the direct correspondence between this and the projection step in the derivation through Galerkin method -- the equation \ref{efixpoint} is essentially the limiting version of the sample-based equation \ref{s-eb}.

\subsection{The geometry of instrumental variables}
\label{s-op}
\begin{figure}[t]
\centering
\begin{minipage}[b]{0.4\linewidth}

\begin{tikzpicture}[line cap=round,line join=round,>=triangle 45,x=0.44cm,y=0.44cm]
\clip(-1.54,-1.8) rectangle (12.86,7.53);
\draw[fill=black,fill opacity=0.1] (7.76,4.61) -- (7.3,4.88) -- (7.03,4.42) -- (7.49,4.15) -- cycle; 
\draw [->] (0,0) -- (7.18,3.64);
\draw [dash pattern=on 3pt off 3pt,domain=-1.54:10.86] plot(\x,{(--18.2-3.64*\x)/-2.18});
\draw [->] (0,0) -- (5,0);
\draw [->] (7.49,4.15) -- (4.84,5.73);
\draw [domain=-1.54:12.86] plot(\x,{(-0-0*\x)/-5});
\draw[color=black] (3.52,2.24) node {$R$};
\draw[color=black] (1.72,-0.80) node {$(\qL)\tilde{V}$};
\draw[color=black] (6.13,6.03) node {$\text{Range}{(\Xi \Phi)}$};
\draw[color=black] (9.0,-0.80) node {$\text{Range}((I - \gamma P) \Phi)$};
\end{tikzpicture}
\end{minipage}
\hspace{45pt}
\begin{minipage}[b]{0.35\linewidth}
\begin{tikzpicture}[line cap=round,line join=round,>=triangle 45,x=0.7cm,y=0.62cm]
\clip(1.07,-1) rectangle (9.3,4.74);
\draw[fill=black,fill opacity=0.1] (4.4,0) -- (4.4,0.4) -- (4,0.4) -- (4,0) -- cycle; 
\draw [dash pattern=on 3pt off 3pt,domain=1.07:9.3] plot(\x,{(-0-0*\x)/2});
\draw (2,0)-- (4,3);
\draw (4,3)-- (4,0);
\draw (6.04,0.78) node[anchor=north west] {$\text{Range}{(\Phi)}$};
\draw [->] (2,0) -- (2.66,0);
\draw [->] (4,0) -- (3.33,0);
\draw (4,0.09) node[anchor=north west] {$\Pi T\tilde{V}$};
\draw (1.72,0.09) node[anchor=north west] {$\tilde{V}$};
\draw (4.04,3.46) node[anchor=north west] {$T\tilde{V}$};
\end{tikzpicture}
\vspace{-12pt}
\vspace*{16pt}
\end{minipage}
\caption{LSTD can be interpreted as an oblique projection (left) and as a fixpoint algorithm (right).}
\label{fig-oblique}
\end{figure}
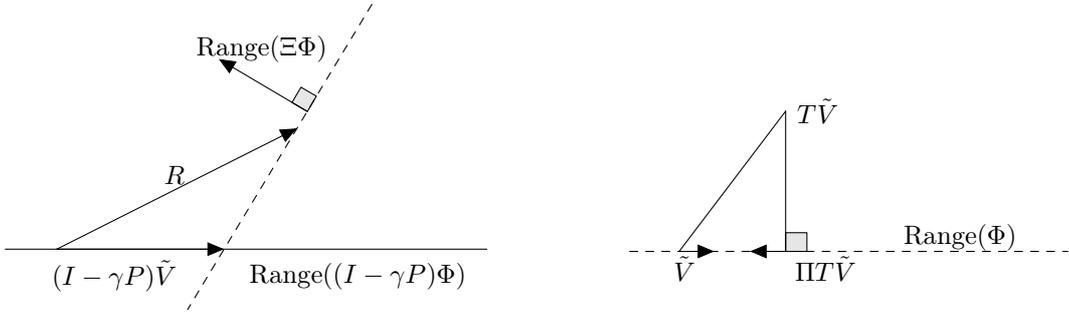

There is one more way to interpret the instrumental variable approach. Observe that the equation $\hat{\Pi}\hat{r} = \hat{\Pi}{D\hat{\Phi}}\hat{w}$, can be rewritten as $\hat{\Pi}({D\hat{\Phi}}\hat{w} - \hat{r}) = 0$. Thus we have that applying the projection amounts to solving $\hat{r} = {D\hat{\Phi}}\hat{w}$ under the constraint that the projection of the residual on the feature space is zero. Therefore LSTD yields the same solution as applying the oblique projection of the rewards on the difference between values of successive states (i.e.\ $D\hat{\Phi}$), along the subspace orthogonal to the column space of $\hat{\Phi}$ (which is the left null-space of $\hat{\Phi}$). See also figure \ref{fig-oblique}.

Recall the formula for the coefficients of the oblique projection on the columns space of $X$ orthogonal to the column space of $Y$, which is $X (Y^\top X)^{-1}Y^\top$. The corresponding generalized pseudoinverse of $X$ is $ (Y^\top X)^{-1}Y^\top$. It is easy to verify that putting $X = (\qL)\Phi$ and $Y = \Xi \Phi$ into $ (Y^\top X)^{-1}Y^\top$ recovers the LSTD solution. Notice that in this case, the projected vector, $X (Y^\top X)^{-1}Y^\top$ corresponds to obtaining the `smoothed rewards' corresponding to the approximate value function (i.e.\ $(\qL)\tilde{V}D$, or what the rewards would have been if there had been no approximation of the value function). Now there is also a different way of defining the projection, namely we can project not the reward vector but the true value function \cite{scherrer-brm}. In this case, setting $X = \Phi$ and $Y = (\qL)^\top \Xi \Phi$ again produces the LSTD solution $w$ (note that now, we are projecting the true value function, not the rewards). Notice that in this case the projected vector corresponds to the approximate value function.

Notice that formally speaking, in both the interpretation as a projection of the reward vector and the value function, we also need another condition to call LSTD an oblique projection -- in order for the formula $X(Y^\top X)^{-1}Y^\top$ to mean a projection on $\text{Range}(X)$ orthogonal to $\text{Range}(Y)$, we need the condition that the orthogonal complement of $\text{Range}(X)$ and $\text{Range}(Y)$ should be complementary subspaces. We will now claim that this is the case in either of the above ways of thinking about LSTD as a projection. To do this, we will prove the following statement. We denote by $k$ the number of columns in $\Phi$ (they are known to be linearly independent by assumption \ref{assm-1}).

\begin{lemma}
For any invertible matrices $A$, $B$, and $\Phi$ is of full column rank, we have the following equivalence.
\[
\text{Range}(A\Phi)^\perp \oplus \text{Range}(B\Phi) = \mathbb{R}^n \quad \Leftrightarrow \quad \neg \exists z. \Phi^\top A^\top B\Phi z = 0 
\] 
\end{lemma}
\begin{proof}
First, we note that the dimension of $\text{Range}(B\Phi)$ is $k$ since $B\Phi$ is full column rank and the dimension of $\text{Range}(A\Phi)^\perp$ is exactly $n-k$ since $A$ is invertible. The argument in the left-to-right direction is as follows: if $\exists z. \Phi^\top A^\top B\Phi z = 0$, then there would be a vector, $B\Phi z$, which is both in $\text{Range}(B\Phi)$ and $\text{Range}(A\Phi)^\perp$. Therefore these two subspaces cannot sum to the $n$-dimensional space if they share a common vector. This contradiction finishes the argument. The argument in the right-to-left direction is thus: there is no non-zero vector in both $\text{Range}(B\Phi)$ and $\text{Range}(A\Phi)^\perp$, then because of their dimensions they have to sum to the whole space $\mathbb{R}^n$. 
\end{proof}

We now see that the condition $\neg \exists z. \Phi^\top A^\top B\Phi z = 0$ is fulfilled in the case of LSTD because by assumption \ref{assm-1} the matrix $\Phi^\top A^\top B\Phi$, and hence also $\Phi^\top B^\top A\Phi$ has to be invertible. In this expression, we can substitute $A = I$ and $B = (\qL)^\top \Xi$ or alternatively $A = \qL$ and $B = \Xi$ to obtain either of the interpretations of LSTD as projection outlined above. We note that in either case, $B\Phi$ is full column rank by assumption \ref{assm-1} together with the fact in appendix \ref{sasm} and $A$ is invertible since $P$ is a Markov matrix. 

\subsection{Connection with the iterative TD algorithm}
\label{td-iterative}
We have seen in section \ref{sec-iv} that the equality $\ex{\phi^\top e_{s}} = -\Phi^\top\Xi R + \Phi^\top\Xi(I - \gamma P)\Phi w = 0$ is crucial for the development of the algorithm and indeed equivalent to the obtained estimator for $w$ (equation \ref{galerkinresult}). We will now show another way of obtaining this equality -- actually, it may be taken do be the \emph{definition} of the algorithm, and used as a justification for the formula \ref{galerkinresult} that stands on its own. We now give the interpretation of this equation is in terms of the iterative TD algorithm \cite{sutton-barto}. We note that the equality $0 = \ex{\phi^\top e_{s}}$ corresponds to saying that the LSTD solution corresponds to the fixpoint of iterative TD, i.e. the point where the expected update is zero. 

Consider now the definition of the iterative TD algorithm \cite{sutton-barto}. We assume for the moment that we have an oracle $V_o$ for the value function and are interested in iteratively solving the optimization problem $\min_w (V_{o}(s) - \valfa{s})^2$ using the approcimation architecture $\valfa{s} = {\phi_s}w$. The iterative update is given by $\nabla_{w} (V_{o}(s) - \valfa{s})^2 = 2 \nabla_{w} \valfa{s} (V_{o}(s) - \valfa{s})$. We now have the following formula for the iteration.

\[\Delta w \propto \underbrace{\nabla_{w} \valfa{s}}_{\phi(s_t)^\top} (\underbrace{\underbrace{(r_{t+1} + \gamma \valfa{s_{t+1}}}_{\text{oracle for value}}) - \valfa{s_t}}_{\text{TD error $ e_{s}$}})\]

Now we have that the update $\Delta w$ at time $t$, is $ \phi(s_t)^\top e_{s_t}$. Setting the expectation of this update to zero gives the desired formula. We also note that the relation between the TD iteration and the LSTD algorithm resembles the chicken-and-egg problem -- one can either, as we did above, consider the iteration a priori knowledge and use that to justify the LSTD fixpoint, or one can start with the fixpoint and treat the iteration as a way of reaching it, motivated by stochastic optimization. LSTD can also be extended to compute the fixpoints of TD($\lambda$) or, more generally other similar algorithms with different traces. For details, see \cite{lstd-lambda-t} in slightly different notation.

\subsection{LSTD as minimization of a quadratic form}
\label{sec-qf}
This section is based on \cite{ICML2011Sun}. It interprets LSTD as the minimization of a quadratic form in the error between the true value function $V{(\cdot)}$ and the approximated value function ${\Phi}w$. We begin by reformulating the formula for the estimator obtained above.
\begin{align*}
w &= \left(\Phi^\top\Xi(I - \gamma P)\Phi\right)^{-1} \Phi^\top \Xi R = \\
   &= \left(\Phi^\top(\qL)^\top \Xi \Phi (\Phi^\top \Xi \Phi)^{-1} \Phi^\top\Xi(I - \gamma P)\Phi\right)^{-1} \\ & \quad \quad \quad \quad \Phi^\top(\qL)^\top \Xi \Phi (\Phi^\top \Xi \Phi)^{-1} \Phi^\top \Xi (\qL) V
\end{align*}
This equality holds because $R = (\qL)V$ and because the matrices $\Phi^\top(\qL)^\top \Xi \Phi$ and $\Phi^\top \Xi \Phi$ are invertible by assumption given by equation  \ref{assm-1}. Now, we introduce the matrix $K$, as below. 
\[
K = (\qL)^\top \Xi \Phi (\Phi^\top \Xi \Phi)^{-1} \Phi^\top\Xi(\qL) = (\qL)^\top \Pi^\top \Xi \Pi (\qL)
\]
We note that $\Xi \Phi (\Phi^\top \Xi \Phi)^{-1} \Phi^\top\Xi = \Xi\Pi = \Pi^\top \Xi = \Pi^\top \Xi \Pi$, where the last equality follows by substituting the definition of $\Pi$ and cancelling the inverted term. Therefore we have $w = \left( \Phi^\top K \Phi \right)^{-1} \Phi^\top K V$. But this is the solution to the well-known optimization problem: $w = \argmin_{w'} \| V - \Phi w' \|_{K} = \argmin_{w'} (V - \Phi w')^\top K (V - \Phi w')$. Thus we gain an insight about approximation $\valfa{\cdot}$ of equation \ref{eb} -- instead of minimizing the norm $\| \cdot \|_{\Xi}$, which would yield us $\bar{V}$, we minimize the different norm $\| \cdot \|_K$, thus gaining the ability of efficiently estimating the solution from samples. Note that we can also repeat the above reasoning, without the multiplication by $(\Phi^\top \Xi \Phi)^{-1}$, to obtain the matrix $K' = (\qL)^\top \Xi \Phi \Phi^\top\Xi(\qL)$ which also defines a valid minimization -- this is the way the equivalence was originally introduced in \cite{schoknecht2002optimality}.

\subsection{LSTD is a subspace algorithm}
\label{sec-sa}
In section \ref{s-op}, we have shown that the algorithm can be thought of as an oblique projection along the subspace orthogonal to the feature space. Here, we make explicit the property that LSTD only depends on the features through the subspace they span i.e. any full-rank transformation (i.e.\ basis change) $C$ of features does not influence the value function. To see this, consider the sample estimate we derived in earlier sections, where we use the transformed features $\hat{\Phi} C$ instead of $\hat{\Phi}$.
\begin{align*}
\hat{V}_C &= \hat{\Phi} C \hat{w}_C = \hat{\Phi} C (C^\top \hat{\Phi}^\top D \hat{\Phi} C)^{-1} C^\top \hat{\Phi}^\top \hat{r} = \\ &= \hat{\Phi} C C^{-1}(\hat{\Phi}^\top D \hat{\Phi})^{-1} {C^\top}^{-1} C^\top \hat{\Phi}^\top \hat{r} = \hat{\Phi} (\hat{\Phi}^\top D \hat{\Phi})^{-1} \hat{\Phi}^\top \hat{r} = \hat{V}
\end{align*}
As a corollary, we state that LSTD is independent of any scaling of features.

\section{LSTD vs Bellman Residual Minimization}
\subsection{A Decompositions of the LSTD loss}
We now present an interpretation of the minimization defined in equation \ref{eoptgalerkin}, after \cite{antosbrm}. We recall that the minimization in equation \ref{eoptgalerkin} can be rewritten in the following way $\Phi y^\star = \argmin_{y} \| T \Phi y^\star - \Phi y \|_\Xi = \Pi (T(\Phi y^\star))$. Therefore $\Phi y^\star - \Pi (T(\Phi y^\star)) = 0$, or $\|\Phi y^\star - \Pi (T(\Phi y^\star)) \|_\Xi = 0$. Therefore LSTD can be seen to be equivalent to the following optimization problem.
\begin{align}
y^\star = \argmin_{y} \|\Phi y - \Pi (T(\Phi y)) \|_\Xi
\label{lstd-normzero}
\end{align}

We note that this expression has no recursion and that the minimization is guaranteed to reach the optimum value of zero. We can now rewrite the norm as follows $\|\Phi y - \Pi (T(\Phi y)) \|_\Xi = \|\Phi y - T(\Phi y) \|_\Xi - \| \Pi (T(\Phi y)) - T(\Phi y)\|_\Xi$, where the equality follows from the Pythagorean theorem and the fact that $\Phi y - \Pi (T(\Phi y))$ and $\Pi (T(\Phi y)) - T(\Phi y)$ are orthogonal vectors, with respect to the $\Xi$-weighted inner product, which corresponds to $\Pi$. We thus obtain the following formula for the LSTD solution.
\begin{align}
y^\star = \argmin_{y} \|  \! \underbrace{\Phi y - T(\Phi y) }_{\text{Bellman residual}} \!\|_\Xi - \| \Pi (T(\Phi y)) - T(\Phi y)\|_\Xi
\label{eoptbrm}
\end{align}
We see that the LSTD algorithm minimizes a quantity which is the Bellman residual minus the reprojection error on the feature space. We discuss in section \ref{sec-brm} the difference between simply minimizing the Bellman residual only and the LSTD algorithm.    

Another way to interpret the LSTD loss is to see it as a nested optimization problem \cite{geist-l1-br}, which leads to the following two equivalent formulations. First, define the projection in the following way. 
\begin{align}
h^\star(y) = \argmin_{h} \| \Phi h - T(\Phi y)\|_\Xi
\label{eopt-tfeature}
\end{align}

Then we plug this for the definition of $\Pi (T(\Phi y))$ in equations \ref{lstd-normzero} and \ref{eoptbrm} respectively, giving the following equivalent equations. 
\begin{gather}
y^\star =  \argmin_{y} \|\Phi y - \Phi h^\star(y) \|_\Xi \quad \text{or} \nonumber \\  y^\star = \argmin_{y} \left( \vphantom{\int} \|\Phi y - T(\Phi y) \|_\Xi - \| \Phi h^\star(y) - T(\Phi y)\|_\Xi \right) 
\label{eopt-fspace} 
\end{gather}

\subsection{Comparison with BRM loss}
\label{sec-brm}
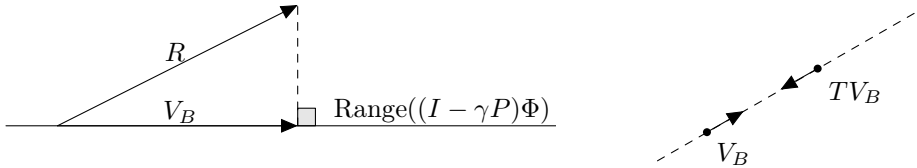
\begin{figure}[t]
\centering
\hspace{20pt}
\begin{minipage}[b]{0.4\linewidth}
\begin{tikzpicture}[line cap=round,line join=round,>=triangle 45,x=0.44cm,y=0.44cm]
\clip(-1.54,-0.67) rectangle (14.9,4.15);
\draw[fill=black,fill opacity=0.1] (7.71,0) -- (7.71,0.53) -- (7.18,0.53) -- (7.18,0) -- cycle; 
\draw [->] (0,0) -- (7.18,3.64);
\draw [->] (0,0) -- (7.18,0);
\draw [domain=-1.54:14.9] plot(\x,{(-0-0*\x)/-7.18});
\draw (7.98,1.16) node[anchor=north west] {$\text{Range}{((\qL) \Phi})$};
\draw [dash pattern=on 3pt off 3pt] (7.18,0)-- (7.18,3.67);
\draw[color=black] (3.52,2.24) node {$R$};
\draw[color=black] (3.67,0.41) node {$V_B$};
\end{tikzpicture}
\end{minipage}
\hspace{55pt}
\begin{minipage}[b]{0.35\linewidth}
\begin{tikzpicture}[line cap=round,line join=round,>=triangle 45,x=0.7cm,y=0.62cm]
\clip(1.07,-1.02) rectangle (5.98,2.74);
\draw [dash pattern=on 3pt off 3pt,domain=1.07:5.98] plot(\x,{(-2.73--1.36*\x)/2.08});
\draw [->] (2,0) -- (2.69,0.45);
\draw [->] (4.08,1.36) -- (3.38,0.91);
\draw (2,-0.03) node[anchor=north west] {$V_B$};
\fill [color=black] (2,0) circle (1.5pt);
\fill [color=black] (4.08,1.36) circle (1.5pt);
\draw[color=black] (4.79,0.85) node {$TV_B$};
\end{tikzpicture}
\vspace*{0pt}
\end{minipage}
\caption{BRM as projection of rewards (left) and minimizing the Bellman residual (right). Cmp. fig. \ref{fig-oblique}}
\label{fig-brm}
\end{figure}

Instead of constructing the oblique projection as described in the previous sections, we can use a simpler algorithm, known as the Bellman Residual Minimization, which corresponds directly to projecting the rewards on the differences between successive states (see figure \ref{fig-brm}) -- i.e.\ it is similar to LSTD except the projection is orthogonal, not oblique. BRM can be interpreted as the un-nested version of the optimization from the previous section.
\begin{align}
g^\star = \argmin_{g} \| \Phi g - T(\Phi g)\|
\end{align} 
The reason LSTD was originally introduced as an improvement over BRM \cite{lstd-bb} is that for BRM, we do not have a justification in terms of a statistical model similar to the one we had in section \ref{sec-iv} -- the noise terms are correlated, so we cannot use a similar reasoning to claim consistency of BRM. But of course the fact that one line of deriving an algorithm doesn't work for BRM does not mean that the algorithm is wrong -- there may be other justifications available. Interestingly, it can be shown that under our assumption \ref{assm-1} the two approaches are similar (the argument comes from chapter 4 of \cite{pires-msc}). Indeed, we have from the previous section (compare equation \ref{lstd-normzero}) that LSTD is similar except for the presence of the projection $\Pi$. It is sometimes useful to have formulas that make the difference between the two algorithms explicit in different formulations of each algorithm. The algebraic relationships between the two algorithms are summarized in the table below.

\vspace{\baselineskip }
\noindent\begin{tabularx}{\textwidth}{lcr}
\hline
LSTD &\hspace*{8.5em}&  BRM \\
\hline
$\min_{w} \| \Pi T  {\Phi}w - {\Phi}w\|_\Xi $ && $\min_{w} \| T  {\Phi}w - {\Phi}w\|_\Xi $ \\
$\min_{w} \| T\Phi w - \Phi w \|_\Xi - \| \Pi T\Phi w - T\Phi w\|_\Xi$ && $\min_{w} \| T\Phi w - \Phi w\|_\Xi$ \\
$w = \left(\Phi^\top\Xi L \Phi\right)^{-1} \Phi^\top \Xi R \text{,}\quad L=I - \gamma P $ 
&& $w = \left(\Phi^\top L^\top\Xi L\Phi\right)^{-1} \Phi^\top L^\top \Xi R$\\
$\min_{w'} \| V - \Phi w' \|_{(\qL)^\top \Pi^\top \Xi \Pi (\qL)}$ && $\min_{w'} \| V - \Phi w' \|_{(\qL)^\top \Xi (\qL)}$\\
${\Phi}w = \Pi T {\Phi}w$ && ${\Phi}w = \underbrace{\Phi(\Phi^\top L^\top \Xi \Phi)^{-1} \Phi^\top L^\top \Xi}_{\text{oblique projection, see \cite{scherrer-brm}}}  T {\Phi}w$ \\ 
\hline
\end{tabularx}
\vspace{\baselineskip }

There has been renewed interest in the analysis of the difference between the two algorithms. One argument \cite{scherrer-brm} is that in an off-line setting (i.e.\ in the situation when the weighing coefficients are different from the stationary distribution of the MRP, a scenario we do not consider in this paper) a performance bound can be shown about BRM that is impossible to derive about LSTD \cite{scherrer-brm}; on the other hand LSTD remains widely used in practice.  

There is yet one more feature that means that LSTD is preferable to BRM is some practical cases -- while with LSTD, as we have shown above, we only need one sequence of samples of features of states and a sequence of samples of reward to obtain an estimate of the value function; but with BRM we need to have two samples of the features of states.

\subsection{Sample estimate of the BRM value function}
We will now show a way to obtain a sample-based estimate $\hat{w}_B$ of the BRM solution, based on section 3.1 of \cite{maillard2010finite}. We want to minimize the expectation $\ex{({\phi}w_B - {\phi'}w_B - r)^2}$. We have the sampled features $\hat{\Phi}^1$ and the sampled rewards $\hat{r}$. We also have a second set of sampled features $\hat{\Phi}^2$. The sampled features are produced using the following process: given the trajectory $s_1, s_2, \dots$, the features in $\hat{\Phi}^1$ are $\phi(s_1), \phi(s_2), \dots$ while the features in $\hat{\Phi}^2$ correspond to `alternative' states $s_2', s_3', \dots$ sampled from $P(\cdot|s_1), P(\cdot|s_2),\dots$. In other words, the features in $\hat{\Phi}^2$ describe where the MRP might also have gone to given a particular previous state. Of course, such sampling is only possible if we have a model of the transition dynamics of the MRP. Now, we can write a sample-based approximation to the expectation given above as $\hat{E} = \frac1{N-1} \sum_{i=1}^{N-1}({\hat{\Phi}^1(i)}w_B - \gamma {\hat{\Phi}^1(i+1)}w_B - \hat{r}(i))({\hat{\Phi}^1(i)}w_B - \gamma {\hat{\Phi}^2(i)}w_B - \hat{r}(i))$, where the notation $\hat{\Phi}^1(i)$ means selecting row $i$ of the matrix $\hat{\Phi}^1(i)$ (i.e.\ the $i$-th feature in the trajectory). We can now introduce the notation $\hat{\Psi}^1 = \hat{\Phi}^1(1:N-1) - \gamma \hat{\Phi}^1(2:N)$ and $\hat{\Psi}^2 = \hat{\Phi}^1(1:N-1) - \gamma \hat{\Phi}^2(1:N)$, where the colon notation denotes ranges of rows. With this notation, we have that $w_B^\top ({\hat{\Psi}^1})^\top \hat{\Psi}^2 w_B = w_B^\top ({\hat{\Psi}^2})^\top \hat{\Psi}^1 w_B = \sum_{i=1}^{N-1} ({\hat{\Phi}^1(i)}w_B - \gamma {\hat{\Phi}^1(i+1)}w_B)({\hat{\Phi}^1(i)}w_B - \gamma {\hat{\Phi}^2(i)}w_B)$. It can now be seen after a few rearrangements that $\hat{E} = \frac1{N-1} \left( w_B^\top ({\hat{\Psi}^1})^\top \hat{\Psi}^2 w_B - \hat{r}^\top (\hat{\Psi}^1 + \hat{\Psi}^2)w_B + \hat{r}^\top \hat{r} \right) = \frac1{N-1} \left( \frac12 w_B^\top ( ({\hat{\Psi}^1})^\top \hat{\Psi}^2 + ({\hat{\Psi}^2})^\top \hat{\Psi}^1 )w_B  - \hat{r}^\top (\hat{\Psi}^1 + \hat{\Psi}^2)w_B + \hat{r}^\top \hat{r} \right)$. Taking the gradient with respect to $w_B$ leaves the us with the system $( ({\hat{\Psi}^1})^\top \hat{\Psi}^2 + ({\hat{\Psi}^2})^\top \hat{\Psi}^1 )\hat{w}_B = (\hat{\Psi}^1 + \hat{\Psi}^2)^\top \hat{r}$, where we denoted by $\hat{w}_B$ the sample-based BRM solution.

\section{Regularization}
\label{sec-reg}
To overcome the problem of over-fitting, the standard procedure is to add a regularization term to the proposed algorithm. There are many ways of doing that. 

One way, proposed by \cite{kolter-ng-reg} is to consider the optimization problem of the fixpoint equation \ref{eoptgalerkin}. We can extend it as follows: $\Phi w = \argmin_{w'} \left ( \| R + \gamma P \Phi w  - \Phi w' \|_\Xi + \beta \| w' \| \right )$. Here, $\beta \geq 0$ is an external parameter of the algorithm, $\| \cdot \|_\Xi$ is the weighted norn and $\| \cdot \| $ is the usual $L_2$ norm. This way of regularizing produces the well-known analytic solution $w_R = \left(\Phi^\top\Xi(I - \gamma P)\Phi +\beta I \right)^{-1} \Phi^\top \Xi R$. In the paper \cite{kolter-ng-reg}, a version is also given where the second norm is $L_1$. In this case, because equation \ref{eoptgalerkin} is a fix-point equation, it is not possible to simply plug the problem into the standard LASSO algorithm, and a new algorithm is necessary (see \cite{kolter-ng-reg} for details).

Before we continue, denote the standard $L_2$-regularized solution of a system of equations $Ax = b$ as $\sollt{A}{b}{\beta} = \argmin_x \| Ax - b\|_{\Xi} + \beta \| x \|_2 = (A^\top \Xi A + \beta I)^{-1} A^\top \Xi b $. Denote the version with $L_1$ regularization as $\sollo{A}{b}{\beta} = \argmin_x \| Ax - b\|_{\Xi} + \beta \| x \|_1$ (this has no explicit analytic form as has to be computed using an algorithm, typically LASSO). 

A second way of regularization, introduced in \cite{geist-l1-br} is to add regularization to equation \ref{eopt-fspace}, giving the following optimization problem.
\[
y^\star =  \argmin_{y} \|\Phi y - \Phi h^\star(y) \|_\Xi + \|y\|_{1 \; \text{or} \; 2}
\]
In the above, the latter norm may be either of $L_2$ or $L_1$. A quick calculation shows that this is the same as regularizing the system of equations \ref{egalerkin}. This idea therefore corresponds to the solutions $\soll{ }{\Phi(I - \gamma (\Phi^\top \Xi \Phi)^{-1} \Phi^\top \Xi P \Phi))}{\Phi (\Phi^\top \Xi \Phi)^{-1} \Phi^\top \Xi R}{\beta}$ for each of the discussed norms. 

Another way is adding regularization directly to the equation where we have already solved for $w$, that is, $w = A^{-1} b$, where $A=\left(\Phi^\top\Xi(I - \gamma P)\Phi\right)$ and $b = \Phi^\top \Xi R$. If we regularize with $L_2$, this corresponds to the solutions $\sollt{A}{b}{\beta}$. This (together with other versions, that do not map to LSTD), has been done in \cite{pires-msc}, where the author also derives finite-sample error bounds.  

It is also possible to combine some of the above ways together, after the manner of \cite{hoffman-lstdreg}, and to use other sparsifiers in place of $L_1$. In \cite{geist-lstdreg}, for instance, the Dantzig selector is employed, which leads to a considerable simplification of the optimization problem (the optimization reduces to a linear program). 

A yet different approach \cite{painter2012greedy} to regularization is to keep the algorithm itself unchanged and instead do feature selection beforehand. Even if the feature selection algorithm is very simple (greedy based on correlation with residual), simulations \cite{painter2012greedy} suggest that doing feature selection leads to performance essentially the same as the approaches described above. Because greedy feature selection is so simple, this suggests that regularization of LSTD is not yet really a fully solved problem.

A property of all the above regularizers is that we lose the invariance of the algorithm w.r.t.\ the choice of basis for the feature space, which can be seen as a natural characteristic of LSTD\footnote{Indeed section 4 of \cite{hoffman-lstdreg} deals with how to perform standardization of features before plugging them into optimization.}. It is not clear whether the property would be worth preserving in a regularized version -- sparsity by its very nature is not invariant to transformations of features, even linear ones and there is a general tendency that a more specialized algorithm will have less generic properties.

\section{The Episodic version of LSTD}
\label{sec-episodic}
In the other sections of this paper, we have considered the case where the MRP never terminates and convergence is defined by taking the limit with respect to the length of a trajectory. We are now interested in extending our observations to the case where there is a termination state. The limit will now be the with respect to the number of episodes being accumulated. First, let us note that the formula $w = \ex{\phi^\top (\phi - \gamma \phi')}^{-1} \ex{\phi^\top r_s}$ is still valid in this case. We simply have to give new meaning to the expectation terms. 

We will now start by giving a design-based variant for the algorithm. All transitions in a terminating MRP can be described using a rectangular matrix $P_t$, where the last column is meant to denote termination. We assume in the following that the starting state of the MRP is the first state. We also assume that the matrix $P_t$ is such that the MRP will always eventually terminate. We first need to construct a state distribution $\Xi$. To do this, we append the row $[1,  0 \dots 0]$ to the matrix $P_t$, producing the square matrix $P_a$, which assumes that the MRP restarts after reaching the termination state. Now, the diagonal entries of the matrix $\Xi$ are the entries of the left eigenvector of $P_a$ which corresponds to eigenvalue one. Now we also construct another square matrix, $P$, which we obtain by appending the row $[0 \dots 0, 1]$ to the matrix $P_t$. This matrix assumes that the agent stays in the termination state forever. The intuition behind this is the following: the matrix $P$ describes the true dynamics of the MRP, but in order to have a meaningful state distribution we need to take into account the fact that we have multiple episodes -- hence the definition of the matrix $P_a$, which models restart. Having defined the above matrices, we may use the standard formula in the following way.
\[
\Xi=\diag{\xi} \; \text{such that } \xi P_a = \xi \quad \text{and} \quad w = \left(\Phi^\top \Xi (I - \gamma P)\Phi\right)^{-1} \Phi^\top \Xi R
\] 
Here, we assume that the last feature vector (i.e.\ the one corresponding to the state modelling termination) is zero. By definition, the final element of $R$ is also zero.

It can be seen that the sample-based variant is the same as in the case of one long trajectory, except for the additional summation over the episodes. We note we use here the fact that the termination state has the feature of zero (so that we can still use the matrix $D$ -- there is no subtraction in the last row, but it doesn't matter since the last state is the terminal state). The formula looks as follows, where the sum goes over episodes.
\[ \hat{w} = (\textstyle \sum_e \hat{\Phi}e^\top D_{S_e} \hat{\Phi}e)^{-1} (\textstyle \sum_e \hat{\Phi}e^\top \hat{r}e)\]

\section{Summary of Contributions}
We have provided a detailed survey of the different ways in which LSTD can be obtained. Our derivation of LSTD using instrumental variables, is, to our knowledge, the first one which is correct. We also made explicit and formal an argument concerning the invertability of the matrix that appears in the LSTD solution (see appendix \ref{app-proofs}).  Moreover, we have derived geometric interpretations of the LSTD fixpoint (independently of the work of Scherrer \cite{scherrer-brm}, which we only became aware of afterwards).  We also provided an exhaustive comparison with the BRM algorithm as well as surveyed the methods that can be used to regularize the LSTD solution. Finally, we formally described the episodic version of LSTD, which was already implicitly known before, but not formalized.

\appendix
\section{Proof of a fact about equation \ref{assm-1} for LSTD.}
\label{app-proofs}
\label{sasm}
\begin{lemma}
Assuming $\ex{\phi^\top \phi}$ is invertible, we have that $\ex{\phi^\top (\phi - \gamma \phi_s')}$ is invertible.
\end{lemma}
\begin{proof}
We rewrite the statement in matrix form: $\det(\Phi^\top \Xi \Phi) \neq 0$ implies $\det(\Phi^\top \Xi (I - \gamma P) \Phi) \neq 0$. We will now develop the second expression. By the well-known eigenvalue argument, $I - \gamma P$ is invertible. Assume for the moment $\Xi > 0$ (we will deal with the case when this is not true later). Consider some non-zero vector $x$. We use the assumption to state that $\Phi x \neq 0$ have that $\Phi^\top \Xi (I - \gamma P) \Phi x = 0$ if and only if the vector $y = \Phi x$, which in the column space of $\Phi$ satisfies the condition that $\Xi (I - \gamma P) y$ is orthogonal to the column space of $\Phi$. This implies that $y^\top \Xi (I - \gamma P) y = 0$. This holds if and only if $y^\top \left( \frac12 ( \Xi (I - \gamma P) ) + \frac12 ( \Xi (I - \gamma P) )^\top \right ) y = 0$. Now because the matrix defining this quadratic form is symmetric, and thus diagonalizable and with real eigenvalues, we have that this can only be zero if some of the eigenvalues are nonpositive. We will show that this cannot be the case. Rewrite the matrix $\frac12 ( \Xi (I - \gamma P) ) + \frac12 ( \Xi (I - \gamma P))^\top$ as $\Xi (I - \gamma \frac12(P + \Xi^{-1} P^\top \Xi))$. Now because by definition $\Xi = \diag{\xi}$ where $\xi^\top P = \xi^\top$ , we have that $\Xi^{-1} P^\top \Xi V \ones = V \ones$ (where by $\ones$ we denote the vector of all ones); moreover, $\Xi^{-1} P^\top \Xi$ has positive entries. So it is a Markov matrix. Thus $\frac12(P + \Xi^{-1} P^\top \Xi)$ also is a Markov matrix. Thus, $(I - \gamma \frac12(P + \Xi^{-1} P^\top \Xi))$ has eigenvalues in the positive real half-plane. We also know that the eigenvalues of $\Xi (I - \gamma \frac12(P + \Xi^{-1} P^\top \Xi))$ are non-negative since it is a symmetric graph Laplacian. But we cannot have zero eigenvalues, because it would imply that $(I - \gamma \frac12(P + \Xi^{-1} P^\top \Xi))$ also has zero eigenvalues, which we have shown is impossible. This finishes the proof for $\Xi > 0$. 

Now consider the case when we do not have this, i.e. some of the diagonal entries of $\Xi$ are zero. Intuitively, the fact we prove is now obvious since transient states do not influence the values of the expectations. More formally, we can, without loss of generality assume that the states for which the probability given by the stationary distribution is zero have highest indexes (i.e.\ they occur at the back of matrices $\Xi, P$ and $\Phi$). We introduce the following notations for block minors of matrices $\Xi, P$ and the vector $y$ corresponding to the non-transient and transient states.
\[
\Xi =  \left [
	    \begin{array}{c|c}
		 \Xi^f & 0 \\ \hline
         0 & 0
        \end{array} \right ] \quad
P =  \left [
	    \begin{array}{c|c}
		 P_f & P_{nt} \\ \hline
         P_{tn} & P_{tt}
        \end{array} \right ] \quad
y =  \left [
	    \begin{array}{c}
		 y_f  \\ \hline
         y_t
        \end{array} \right ]
\]
Note that in the above, $P_{nt}$ is has to be the zero matrix -- it corresponds to transitions from non-transient states to transient states. Therefore we have that $\Xi (I - \gamma P) y = 0$ implies $\Xi^f (I - \gamma P_f) y_f = 0$ and thus, by the reasoning for the case without transient states, $y_f$ has to be the zero vector. Therefore we have the fact that $\Xi (I - \gamma P) \Phi x = 0$ implies that we have the following.
\[
\Phi x = y =  \left [
	    \begin{array}{c}
		 0  \\ \hline
         y_t
        \end{array} \right ]
\]
We see that this implies that $\Phi^\top \Xi \Phi x = 0$. But we know from our assumption  $\det(\Phi^\top \Xi \Phi) \neq 0$ that this is only possible for $x = 0$. 
\end{proof}

{
\section*{Bibliography}
\begin{scriptsize}
\renewcommand*{\refname}{ \vspace*{-5mm} }
\bibliography{./ucl-bb}
\bibliographystyle{plain}
\end{scriptsize}
}
\end{document}